\documentclass[runningheads]{llncs}
\usepackage[T1]{fontenc}
\usepackage{graphicx}
\usepackage{makecell}
\usepackage{url}
\usepackage{hyperref}
\usepackage{xcolor}
\usepackage[misc]{ifsym}
\usepackage{color}

\begin{document}
\title{What Streaming Anomaly Detection Finds \\ (and Misses) in Industrial Time Series}
\toctitle{What Streaming Anomaly Detection Finds (and Misses) in Industrial Time Series}
\titlerunning{Streaming Anomaly Detection in Industrial Time Series}
%
\author{Magali Parrino\inst{1,2} {\Letter} \and 
Antoine Ajenjo\inst{1} \and 
Emmanuel Remy\inst{1}\and 
Pierre Stephan\inst{1} \and
Paul Boniol\inst{2}}

\tocauthor{Magali Parrino, Antoine Ajenjo, Emmanuel Remy, Pierre Stephan, Paul Boniol}

\authorrunning{M. Parrino et al.}
%
\institute{EDF R\&D, France, \email{first.name@edf.fr} \and
Inria, ENS, CNRS, PSL, France, \email{first.name@inria.fr}}

\maketitle              
\begin{abstract}
EDF relies on continuous monitoring of its power plants to detect anomalies as soon as they occur. Given the absence of a universally optimal streaming method in unsupervised settings, we compare streaming methods with state-of-the-art TSAD models deployed online on a real nuclear power plant dataset. This work also evaluates Automated Anomaly Detection in a streaming context. Results show higher consistency for online TSAD and strong robustness from ensembling strategies.

\end{abstract}
\section{Introduction} 

EDF, as the main French electric utility company, operates numerous types of power plants (e.g., dams, wind farms and nuclear power stations). 
Such facilities are continuously monitored in order to detect anomalies as soon as possible, to prevent further degradation. 
To avoid developing case-specific supervised algorithms requiring expert knowledge, unsupervised streaming algorithms are of high interest as to facilitate their industrial use on a wide range of plants.

Time Series Anomaly Detection (TSAD) is a widely studied field, with numerous methods developed over time~\cite{Boniol24}. 
Overall, recent works~\cite{Liu25} have offered several takeaways, notably that there is no universal best method.
A similar conclusion is reached in the Streaming TSAD field~\cite{Cao24}, which has focused on developing fast and adaptable outlier detection methods. 
However, concerns were raised~\cite{Parrino26} over the suitability of these methods for real-world applications, as anomalies might manifest as anomalous sub-sequences, which outliers-focused methods struggle to correctly detect. 
Therefore, pursuing a universal unsupervised streaming anomaly detector might be unrealistic. 

Automated Anomaly Detection (AutoAD)~\cite{Bahri22}, comprising of ensembling strategies and model selection, has emerged as a promising solution.
It has, however, not yet been much explored in a streaming context~\cite{Bahri22}.

\noindent In this paper, we assess the relevance of AutoAD on one nuclear power plant monitoring, comparing Streaming and TSAD methods with basic AutoAD strategies.

\section{Our Use-Case: \textit{Nuclear Power Plants Clogging Issue}}

Nuclear power plants require water (from seas or rivers) to cool down their systems. 
The water is pumped through filters that can be clogged by foreign bodies. 
This clogging can prompt a scram (automatic reactor shutdown). 
Each scram involves generation downtime, restart delays 
and specialized teams for onsite intervention. 
Early clogging detection is therefore a high stakes task.

The data labelled by EDF comes from the Blayais nuclear power plants site, comprised of four units of 900MWe each. 
As it is located at the confluence of the Dordogne river, the Garonne river and the Atlantic ocean, studies have shown~\cite{these_macadam} that the main clogging agents are decaying tree leaves carried by the rivers. 
High river flows carry the scraped clogging agents to the power plants' surroundings, and low tide with high tidal coefficients finish routing them to the water intakes. \\
\noindent\textbf{Studied Data:}
The time series studied comes from both the pumping stations' sensors and external environment parameters.
For each reactor $i\in \{1,...,4\}$, the pressure differential after the filter drums ($\Delta P\_i$) and the number of times a threshold $\tau$ was exceeded ($\Delta P>\tau\_i$) are monitored. In addition, we retrieved the wind speed, the water level at 2 different distances from the site, and the river flow of both the Dordogne and the Garonne, resulting in \textbf{a total of 13 dimensions}.
The data spans from April 1996 to June 2021, with one point per hour, generating a \textbf{time series of 223k points}.
In total, 20 events (grouped in 6 periods) that triggered a scram were reported, as shown in Figure~\ref{visu_exp} (a). \\
\noindent\textbf{Objective:}
We aim to evaluate on the use-case described above the real-time performance of Online methods (TSAD literature put in a streaming context) to Streaming ones (from the Streaming TSAD literature).

\section{Bridging TSAD and Streaming Literatures}

We apply 29 methods on the time series, 19 Online methods and 10 Streaming methods (implementations and references are available in the \href{https://github.com/magaliparrino/StrAD}{{\color{blue}StrAD}} repository).\\
\noindent\textbf{Online Methods:}
Assume access to an (unlabelled) initial batch for model training. 
Afterward, internal parameters are fixed, and inference occurs on incoming data without updates.
The 19 TSAD methods used Online balance vetted machine learning models and recent deep learning architectures.\\
\noindent\textbf{Streaming Methods:}
Process the time series sequentially and allow model updates over time.
The 10 Streaming models evaluated in this work cover different update and forgetting strategies proposed in the literature. \\
\noindent\textbf{TSAD Relevance for Streaming:}
A recent work~\cite{Parrino26} shows that \textit{TSAD methods} applied in an \textit{Online context}
perform overall better than \textit{methods from the streaming literature}, even though these have update and forgetting mechanisms. 
Thus, TSAD methods may perform better than the streaming ones, while their slower inference is of no consequence in our use-case, as the stream velocity is low (one hour between each point).
The behavior of Online models over streams with distribution shifts is studied in depth in a companion benchmark~\cite{Parrino26}.

\section{Experimental Evaluation on our Use-Case}

In this study, we use VUS-PR~\cite{Boniol2025} accuracy measure with a left-buffer of 72 points (i.e., 3 days).
We ignore the right-buffer to favor only early detection.
Moreover, we train on the first $25\%$ of the time series,
where the training set contains 2 clusters of anomalies.
Finally, if needed for the model, we apply z-normalization, with the mean and standard deviation computed on the training batch only.

\begin{figure}[t!]
\includegraphics[width=\textwidth]{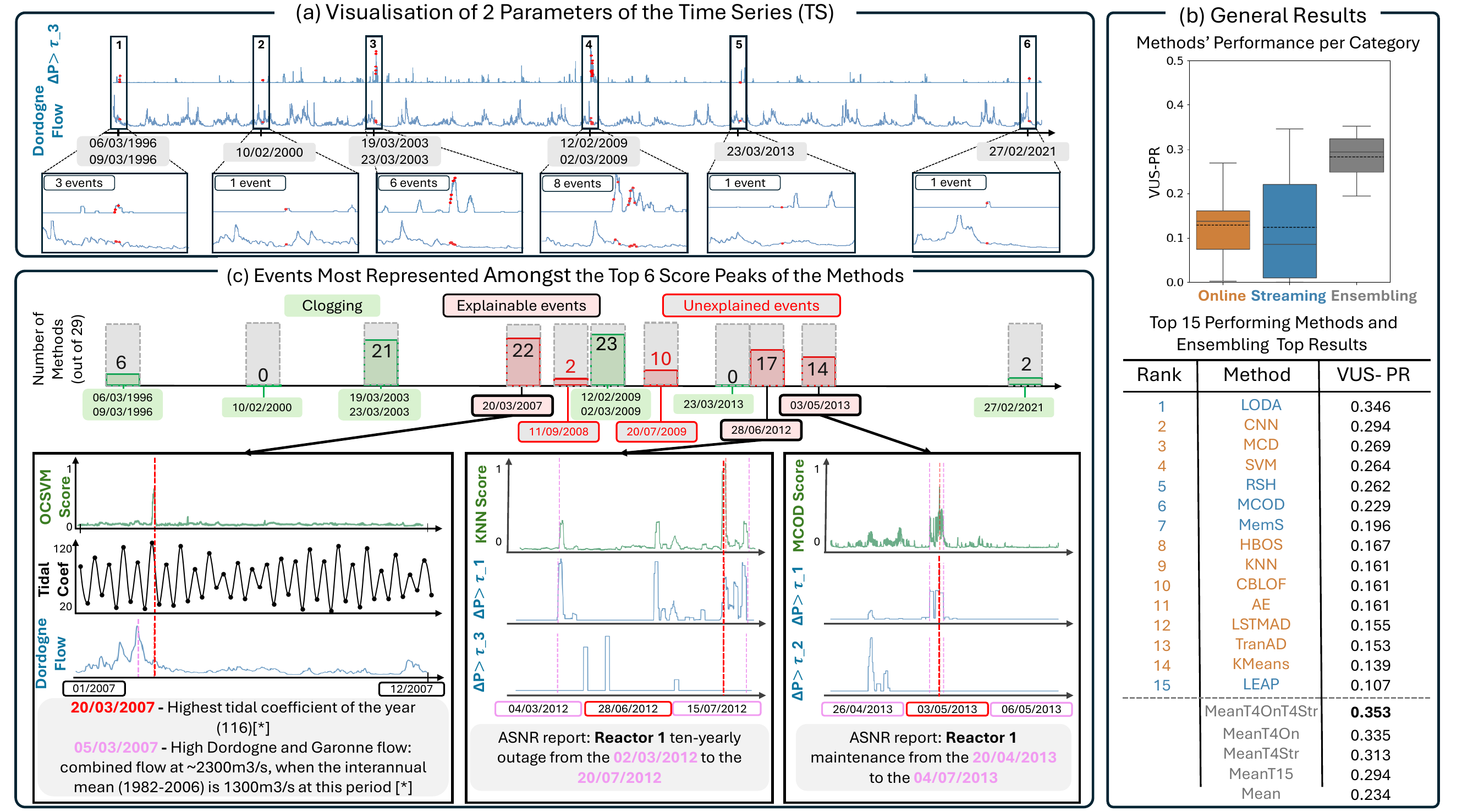}
\caption{(a) Partial Visualisation of the Data; (b) Performances of Online, Streaming, and Ensembling Strategies; (c) Consensus Amongst Top 6 Score Peaks of the 29 Models, Explanation of Some False Positives ([*]: Ifremer data)}
\label{visu_exp}
\end{figure}

\subsection{Online vs. Streaming vs. Ensembling}

An analysis of the distributions presented in Figure~\ref{visu_exp} (b) reveals a notable discrepancy between mean and median performance metrics. 
While the mean performance (represented by the dotted line) is comparable across both categories, the Online median (solid line) is significantly higher than that of the Streaming group.
This statistical divergence is driven by specific outliers: although LODA, a streaming method, achieves the highest individual performance, the Online models demonstrate superior collective consistency.

By aggregating the anomaly scores of the 29 methods, we evaluate ensembling strategies ranging from unsupervised to supervised. Mean denotes the simple average across all 29 models. MeanT\textbf{\textit{k}}\textit{type} denotes the average over the top \textbf{\textit{k}} models within a category (\textit{Online} or \textit{Streaming}); when no type is specified, selection is across the combined pool.
The results underline the robustness of aggregation strategies. 
Unsupervised Mean reaches the top-six ranking, while
basic supervised top-k selections secure top-two ranking and, in one instance, outperform the best-performing individual model.
These findings suggest that AutoAD 
ensures high performance stability through ensemble agreement, as individual model weaknesses are compensated across the pool.

\subsection{Phenomenological Insights}

The advantages of ensembling are further illustrated by the consensus analysis in Figure~\ref{visu_exp} (c). 
Overall, we observe that (1) obvious anomaly clusters (3, 4) are identified by a large majority of models and isolated anomalies (clusters 2, 5, 6) are rarely captured; (2)  certain False Positives (FP) are highly recurrent across the model pool.
Analysis using data from Ifremer (oceanographic research institute) and ASNR (French Nuclear Safety Authority) provides potential explanations for these consensual FP:

\noindent\textbf{20/03/2007 FP}: A high river flow followed by a peak spring tide are highly conducive to clogging. The model consensus suggests an anomaly, though perhaps mitigated by preemptive on-site intervention, preventing a scram.

\noindent\textbf{20/07/2012 and 04/07/2013 FP}: These coincide with scheduled reactor 1 shutdowns. In an unsupervised setting, the resulting variance in operational parameters is flagged by the models as a deviation from the statistical norm.

\section{Conclusion and Implications for Operational Reliability}
Our study highlights two critical considerations for the deployment of \textbf{unsupervised anomaly detection in industrial settings}:
(1) anomaly detectors flag any deviation from a perceived normal behavior, so rare events may be detected alongside the anomaly of interest;
(2) ensembling provides a reliable proxy for the
physical reality of an event, even if exhaustive recall is not achieved. \\
\noindent Therefore, AutoAD is a promising approach toward autonomous monitoring in industrial applications. 
For now, none of the evaluated methods are operationally deployed at EDF.
The primary remaining challenge is the development of unsupervised streaming top-k selection mechanisms.

%
%
%
%

\end{document}